\documentclass{article}

\usepackage[numbers]{natbib}
\usepackage[preprint]{neurips_2022}

\usepackage[dvipsnames]{xcolor}         
\definecolor{linkColor}{rgb}{0.18,0.39,0.62}
\usepackage[utf8]{inputenc} 
\usepackage[T1]{fontenc}    
\usepackage[colorlinks=true,linkcolor=linkColor,citecolor=linkColor,filecolor=linkColor,urlcolor=linkColor]{hyperref}       
\usepackage{url}            
\usepackage{booktabs}       
\usepackage{amsfonts}       
\usepackage{nicefrac}       
\usepackage{microtype}      

\usepackage{graphicx}
\usepackage{arydshln}
\usepackage{booktabs}
\usepackage{multirow}
\usepackage{caption}
\usepackage{subcaption}
\usepackage{makecell}
\usepackage{csquotes}
\usepackage{epigraph}
\usepackage{amsmath}
\usepackage{xcolor}  
\usepackage{cleveref}
\usepackage{pifont}
\usepackage{listings}

\RequirePackage{algorithm}
\RequirePackage{algorithmic}

\usepackage{multirow}
\usepackage{amsmath}
\usepackage{capt-of}
\usepackage{tabularx}
\usepackage{epsfig}
\usepackage{amssymb}
\usepackage{amsfonts}
\usepackage{booktabs}
\usepackage{scalerel}
\usepackage[inline]{enumitem}
\usepackage{listings}
\usepackage{varwidth}
\usepackage[export]{adjustbox}
\usepackage{tikz}
\usetikzlibrary{tikzmark}

\usepackage{stmaryrd}
\usepackage{bbm}
\usepackage{wrapfig}
\usepackage{pifont}

\definecolor{deepblue}{rgb}{0,0,0.5}
\definecolor{officeblue}{RGB}{0,102,204}
\definecolor{deepred}{rgb}{0.6,0,0}
\definecolor{deepgreen}{rgb}{0,0.5,0}
\definecolor{mybrickred}{RGB}{182,50,28}

\definecolor{fillcolor}{RGB}{216,217,252}

\usepackage{etoolbox}
\usepackage{framed}

\newif\ifxetexorluatex
\ifxetex
  \xetexorluatextrue
\else
  \ifluatex
    \xetexorluatextrue
  \else
    \xetexorluatexfalse
  \fi
\fi
\newcommand*\quotesize{60} 
\newcommand*{\openquote}
   {\tikz[remember picture,overlay,xshift=-4ex,yshift=-2.5ex]
   \node (OQ) {\fontsize{\quotesize}{\quotesize}\selectfont``};\kern0pt}

\newcommand*{\closequote}[1]
  {\tikz[remember picture,overlay,xshift=4ex,yshift={#1}]
   \node (CQ) {\fontsize{\quotesize}{\quotesize}\selectfont''};}

\colorlet{shadecolor}{white}

\newcommand*\shadedauthorformat{\emph} 

\newcommand*\authoralign[1]{%
  \if#1l
    \def\authorfill{}\def\quotefill{\hfill}
  \else
    \if#1r
      \def\authorfill{\hfill}\def\quotefill{}
    \else
      \if#1c
        \gdef\authorfill{\hfill}\def\quotefill{\hfill}
      \else\typeout{Invalid option}
      \fi
    \fi
  \fi}
\newenvironment{shadequote}[2][l]%
{\authoralign{#1}
\ifblank{#2}
   {\def\shadequoteauthor{}\def\yshift{-2ex}\def\quotefill{\hfill}}
   {\def\shadequoteauthor{\par\authorfill\shadedauthorformat{#2}}\def\yshift{2ex}}
\begin{snugshade}\begin{quote}\openquote}
{\shadequoteauthor\quotefill\closequote{\yshift}\end{quote}\end{snugshade}}

\usepackage{amsmath,amsfonts,bm}

\def\eqref#1{equation~\ref{#1}}

\def\1{\bm{1}}

\DeclareMathAlphabet{\mathsfit}{\encodingdefault}{\sfdefault}{m}{sl}
\SetMathAlphabet{\mathsfit}{bold}{\encodingdefault}{\sfdefault}{bx}{n}

\newcommand\our{\text{BitNet v2}}
\newcommand\rotbitlinear{\text{$\mathcal{H}$-BitLinear}}
\newcommand\bitnet{\text{BitNet b1.58}}
\newcommand\bitnetx{\text{BitNet a4.8}}

\title{\our{}: Native 4-bit Activations with Hadamard Transformation for 1-bit LLMs}

\author{
 Hongyu Wang\thanks{~Equal contribution. $\diamond$ Corresponding author. S. Ma and F. Wei are with Microsoft Research. H. Wang is with University of Chinese Academy of Sciences.}~~~~Shuming Ma\footnotemark[1]~~~~Furu Wei$^{\diamond}$ \\
{\href{https://aka.ms/GeneralAI}{https://aka.ms/GeneralAI}}
\vspace{-0.4cm}
\\}

\begin{document}
\maketitle
\begin{abstract}
\vspace{-0.2cm}
Efficient deployment of 1-bit Large Language Models (LLMs) is hindered by activation outliers, which complicate quantization to low bit-widths. We introduce \our{}, a novel framework enabling native 4-bit activation quantization for 1-bit LLMs. To tackle outliers in attention and feed-forward network activations, we propose \rotbitlinear{}, a module applying an online Hadamard transformation prior to activation quantization. This transformation smooths sharp activation distributions into more Gaussian-like forms, suitable for low-bit representation. Experiments show \our{} trained from scratch with 8-bit activations matches BitNet b1.58 performance. Crucially, \our{} achieves minimal performance degradation when trained with native 4-bit activations, significantly reducing memory footprint and computational cost for batched inference.

\begin{shadequote}[r]{Richard Feynman}
{There's Plenty of Room at the Bottom.}
\end{shadequote}
\end{abstract}

\vspace{-0.5cm}

\begin{figure}[h]
    \centering
    \begin{subfigure}{\textwidth}
        \includegraphics[width=\textwidth]{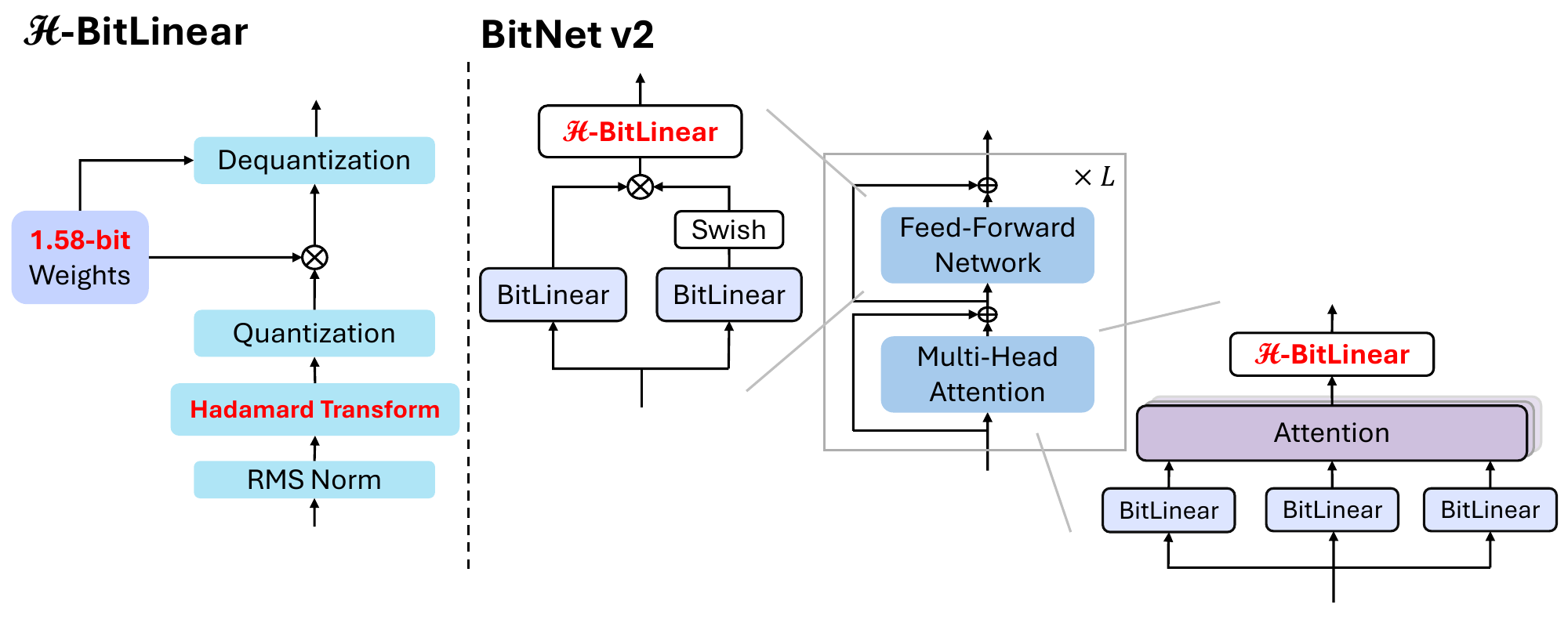}
    \end{subfigure}
    \begin{subfigure}{0.24\textwidth}
        \includegraphics[width=\textwidth]{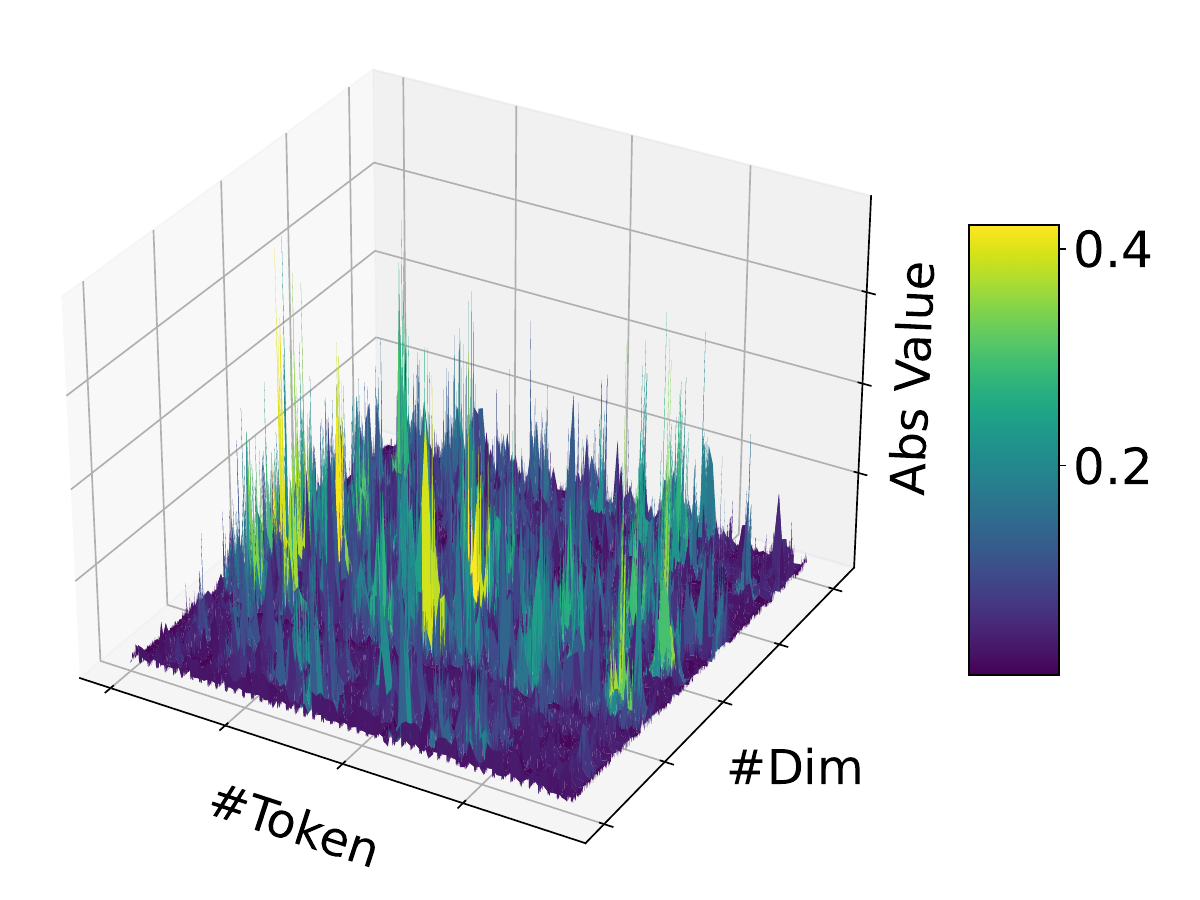}
        \caption{$\mathbf{W}_{\text{o}}$ of BitNet b1.58}
    \end{subfigure}
    \begin{subfigure}{0.24\textwidth}
\vspace{-0.4cm}        \includegraphics[width=\textwidth]{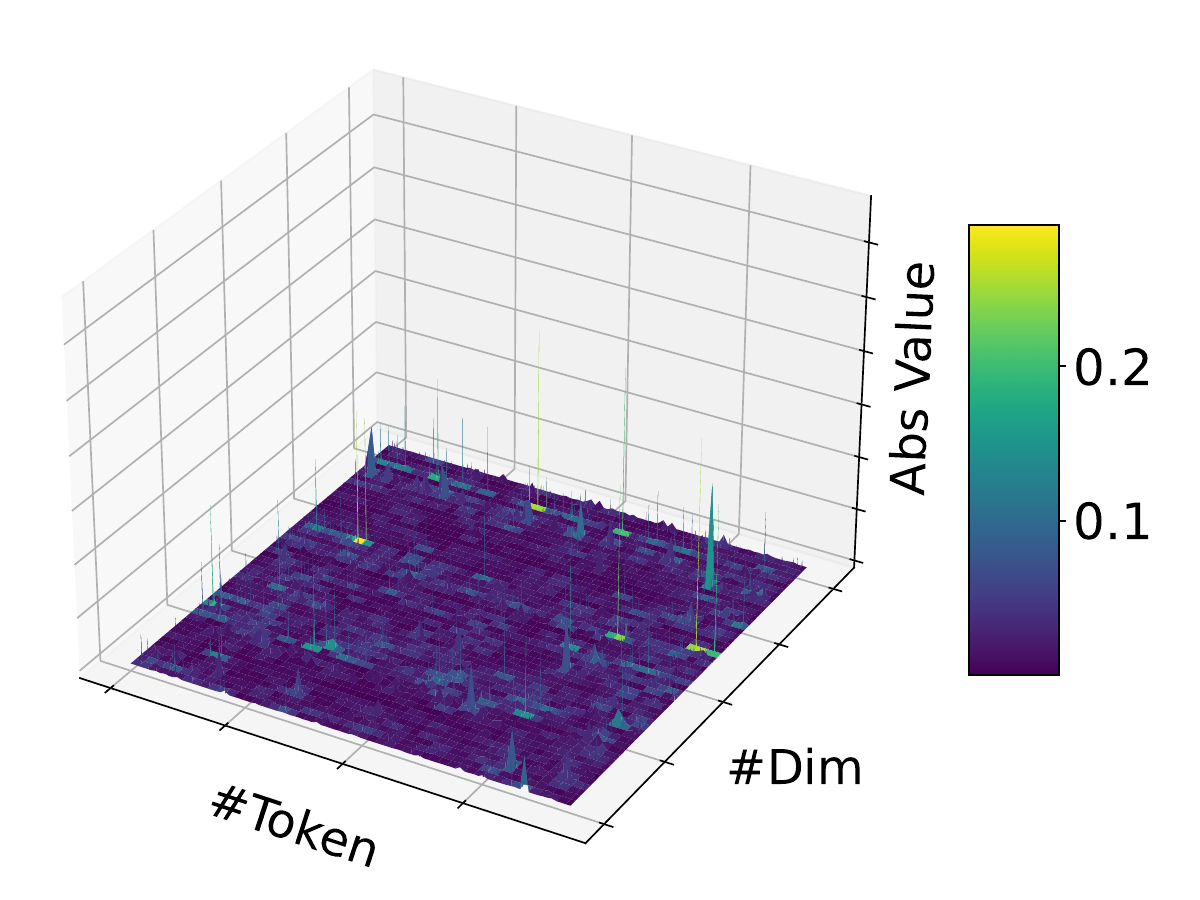}
        \caption{$\mathbf{W}_{\text{down}}$ of BitNet b1.58}
    \end{subfigure}
    \begin{subfigure}{0.24\textwidth}
        \includegraphics[width=\textwidth]{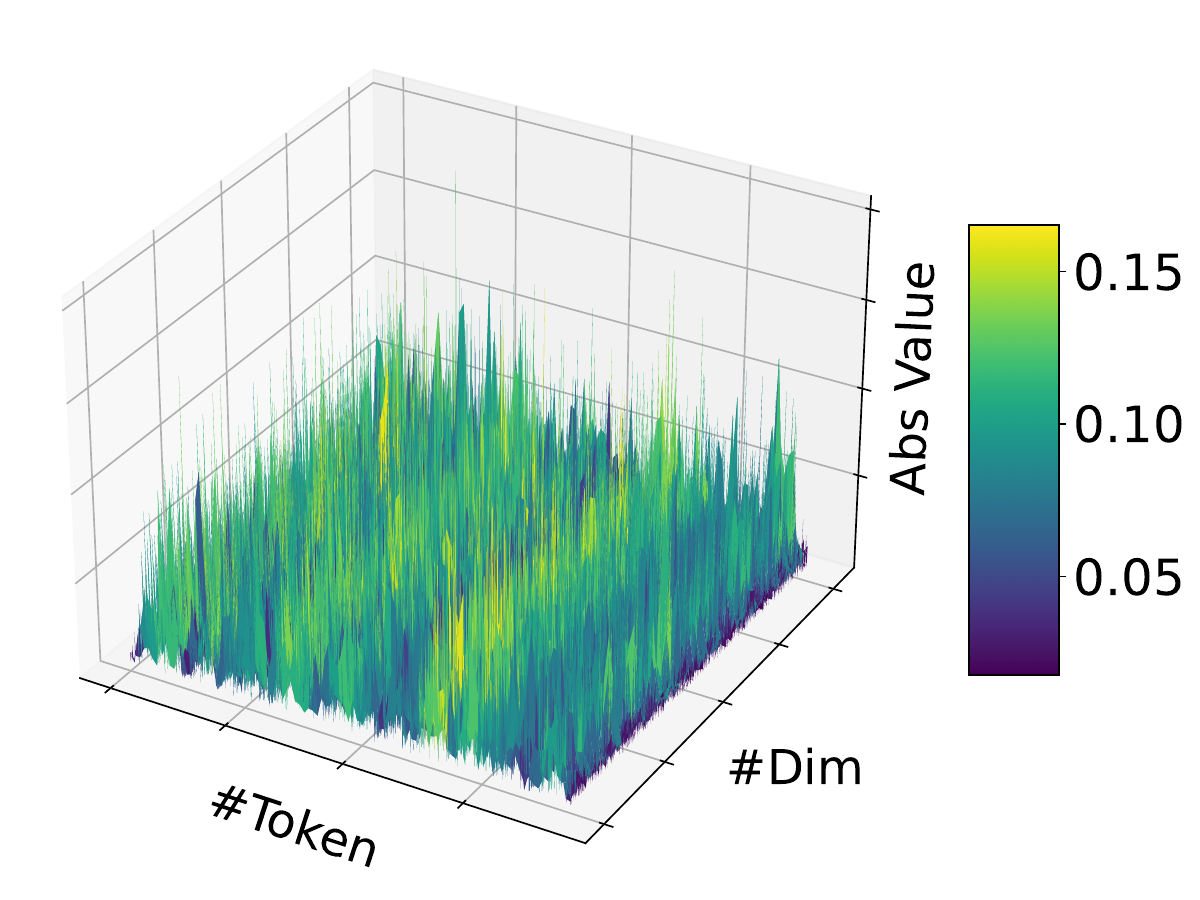}
        \caption{$\mathbf{W}_{\text{o}}$ of \our{}}
    \end{subfigure}
    \begin{subfigure}{0.24\textwidth}
        \includegraphics[width=\textwidth]{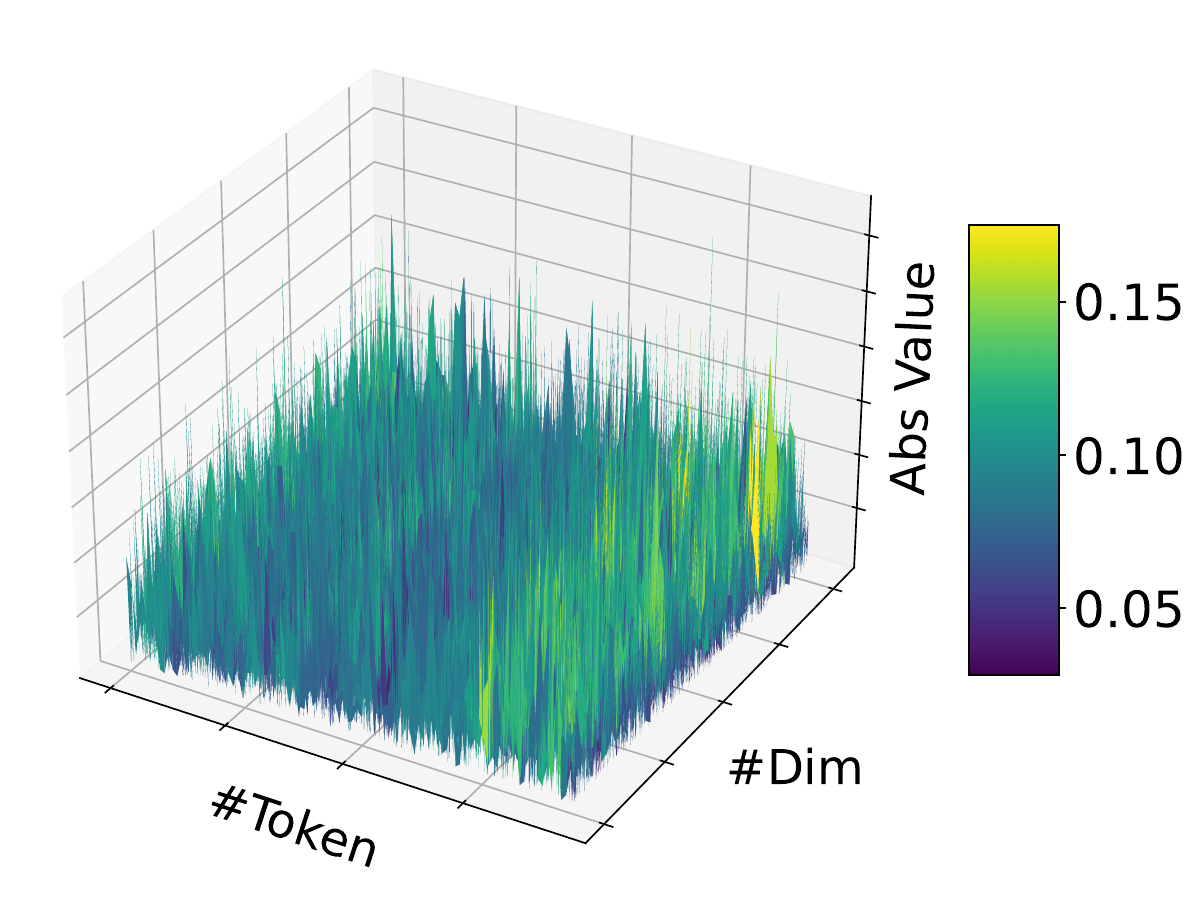}
        \caption{$\mathbf{W}_{\text{down}}$ of \our{}}
    \end{subfigure}  
    \caption{\textbf{Top:} Overview of \our{} and \rotbitlinear{}. \textbf{Bottom:} The distribution of the activation of output projection $\mathbf{W}_{\text{o}}$ in attention and down projection $\mathbf{W}_{\text{down}}$ in FFN. \our{} utilizes \rotbitlinear{} to eliminate the large amount of outlier channels in the intermediate states. The Hadamard transformation reshapes the original sharp distribution into a more Gaussian-like form.}
    \label{fig:overview}
\end{figure}

\newpage


\section{Introduction}

The field of deep learning is rapidly embracing quantization-aware training and low-bit inference, driven by hardware advancements like next-generation GPUs (e.g., GB200) offering native support for 4-bit computations. This promises significant efficiency gains for large-scale models. Pioneering work like BitNet b1.58~\cite{bitnet2} demonstrated that 1.58-bit LLMs can match full-precision performance while drastically reducing inference costs (latency, memory, throughput, energy)~\cite{bitnetcpp}. However, while BitNet b1.58 quantizes weights to 1.58 bits, alleviating memory bandwidth bottlenecks, it retains 8-bit activations. This reliance on 8-bit precision prevents these models from fully leveraging the 4-bit computational capabilities of emerging hardware, shifting the efficiency bottleneck towards computation itself.

Achieving lower bit-width activations is crucial for maximizing hardware utilization, particularly for efficient kernel design in batched inference scenarios. Research~\cite{bitnet48, teal} highlights a key challenge: the non-uniform distribution of activations within LLMs. While inputs to attention and feed-forward network (FFN) layers often exhibit Gaussian-like distributions amenable to quantization, their intermediate states (outputs before final projection) contain significant outliers, hindering aggressive low-bit quantization. BitNet a4.8~\cite{bitnet48} attempted to address this by selectively using 4-bit quantization for inputs and 8-bit sparsification for intermediate states. While achieving minimal performance loss compared to 8-bit activations, sparsification is less suited for maximizing throughput in batched inference, where dense computations are often preferred for hardware efficiency.

To bridge this gap and unlock the full potential of 4-bit computation for 1.58-bit LLMs, we introduce \textbf{\our{}}. Our framework enables \emph{native} 4-bit activations across the model. The core innovation is \textbf{\rotbitlinear{}}, a novel linear layer replacing the standard output projections in attention and down projections in FFNs. \rotbitlinear{} applies an online Hadamard transformation before activation quantization. This strategically reshapes the sharp, outlier-prone distributions of intermediate states into more manageable, Gaussian-like forms, significantly reducing the impact of outliers in 1.58-bit models. We train \our{} from scratch using 8-bit activations, achieving negligible performance loss compared to BitNet b1.58~\cite{bitnet2}. Subsequently, the model can be efficiently fine-tuned with a small amount of data to operate with native 4-bit activations. Extensive experiments demonstrate that our 4-bit \our{} variant achieves performance comparable to BitNet a4.8 while offering superior computational efficiency for batched inference scenarios.

\section{\our{}: Native 4-bit Activations}
We illustrate the architecture of \our{} in Figure~\ref{fig:overview}. We implement \our{} using LLaMA-like components, including RMS normalization~\cite{rmsnorm}, SwishGLU~\cite{swiglu} and removing all bias. Compared to BitNet~\cite{bitnet}, we use \rotbitlinear{} for $\mathbf{W}_\text{o}$ in attention and $\mathbf{W}_\text{down}$ in FFN layers to deal with outlier channels of intermediate states. \our{} is trained with 1.58-bit weights and INT8 activations from scratch, then continue-trained with INT4 activations for all linear layers except input/output embedding.

\subsection{\rotbitlinear{}}
\label{sec:arch}
Following~\cite{bitnet, bitnet2}, as for weight quantization, we use per-tensor \textit{absmean} function to quantize the weights into ternary values, i.e., \{-1, 0, 1\}:

\begin{align}
    &\text{Q}_{w}(\mathbf{W}) = \alpha\text{RoundClip}(\frac{\mathbf{W}}{\alpha+\epsilon}, -1, 1),\,\alpha = \text{mean}(|\mathbf{W}|) \\
    &\text{RoundClip}(X, a, b) = \min(\max(\text{round}(X), a), b)
\end{align}

For low-bit activations, previous works~\cite{bitnet48, teal} have shown that the distributions of the inputs to the attention and feed-forward-network layers (i.e., the activations of $\mathbf{W}_\text{qkv}$ and $\mathbf{W}_\text{up,gate}$) tend to exhibit a Gaussian-like shape, while the intermediates states (i.e., the activations of $\mathbf{W}_\text{o}$ and $\mathbf{W}_\text{down}$) have more outlier channels and massive amount of entries around zero. 

Therefore, we introduce \textbf{\rotbitlinear{}} for $\mathbf{W}_\text{o}$ in attention and $\mathbf{W}_\text{down}$ in FFN layers. \rotbitlinear{} employs a Hadamard transformation before activation quantization to first reduce the number of outlier channels. The Hadamard transformation satisfies that:

\begin{align}
    &\text{Hadamard}(\mathbf{X}) = \mathbf{H_m} \mathbf{X} \\
    & \mathbf{H_m} = \cfrac{1}{\sqrt{2}}\left ( \begin{array}{cc}
       \mathbf{H_{m-1}}  &  \mathbf{H_{m-1}}\\
       \mathbf{H_{m-1}}  &  - \mathbf{H_{m-1}}
    \end{array} \right ),\, \mathbf{H_{0}} = (1)
\end{align}

where $\mathbf{H_m}$ is a $2^m \times 2^m$ matrix and $\mathbf{X} \in \mathbf{R}^n$, $n = 2^m$. We use fast-hadamard-transform\footnote{\url{https://github.com/Dao-AILab/fast-hadamard-transform}} to perform the matrix multiplication, which has $\mathcal{O}(n \log n)$ computational complexity. 

\begin{figure*}[t]
    \centering
    \begin{subfigure}{0.24\textwidth}
        \includegraphics[width=\textwidth]{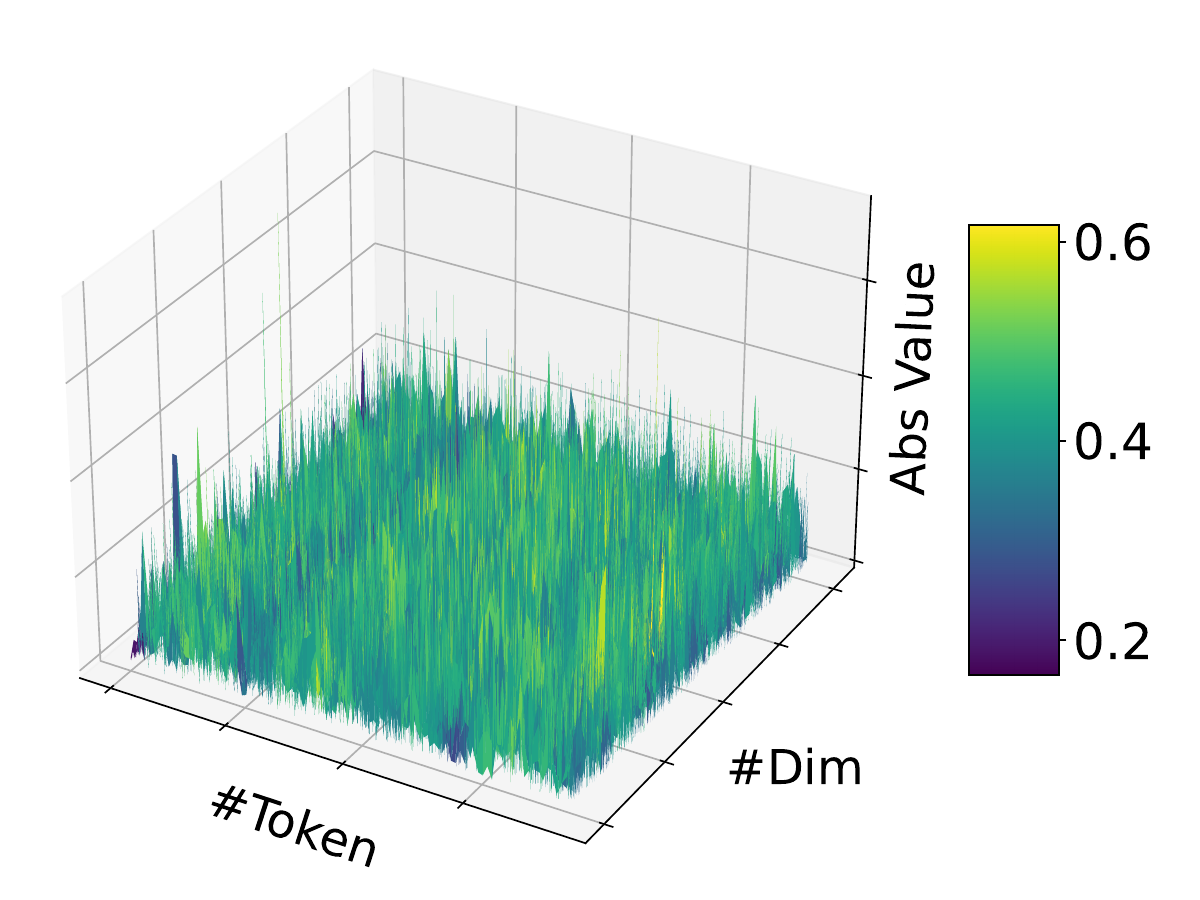}
        \caption{$\mathbf{W}_{\text{qkv}}$ of BitNet b1.58}
    \end{subfigure}
    \begin{subfigure}{0.25\textwidth}
        \includegraphics[width=\textwidth]{pics/3b-a8.layer.15.o.pdf}
        \caption{$\mathbf{W}_{\text{o}}$ of BitNet b1.58}
    \end{subfigure}
    \begin{subfigure}{0.25\textwidth}
        \includegraphics[width=\textwidth]{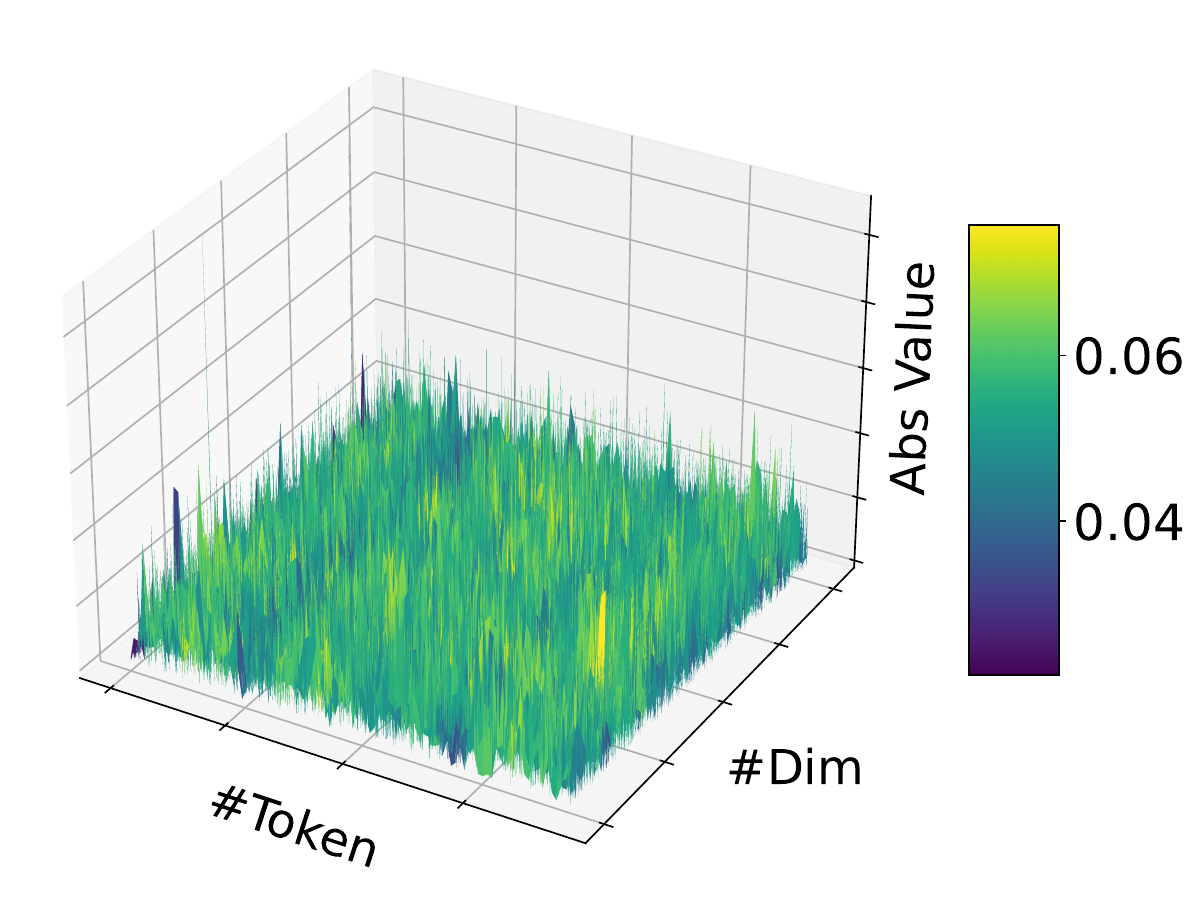}
        \caption{$\mathbf{W}_{\text{up,gate}}$ of BitNet b1.58}
    \end{subfigure}
    \begin{subfigure}{0.24\textwidth}
        \includegraphics[width=\textwidth]{pics/3b-a8.layer.15.down.pdf}
        \caption{$\mathbf{W}_{\text{down}}$ of BitNet b1.58}
    \end{subfigure}
    \begin{subfigure}{0.24\textwidth}
        \includegraphics[width=\textwidth]{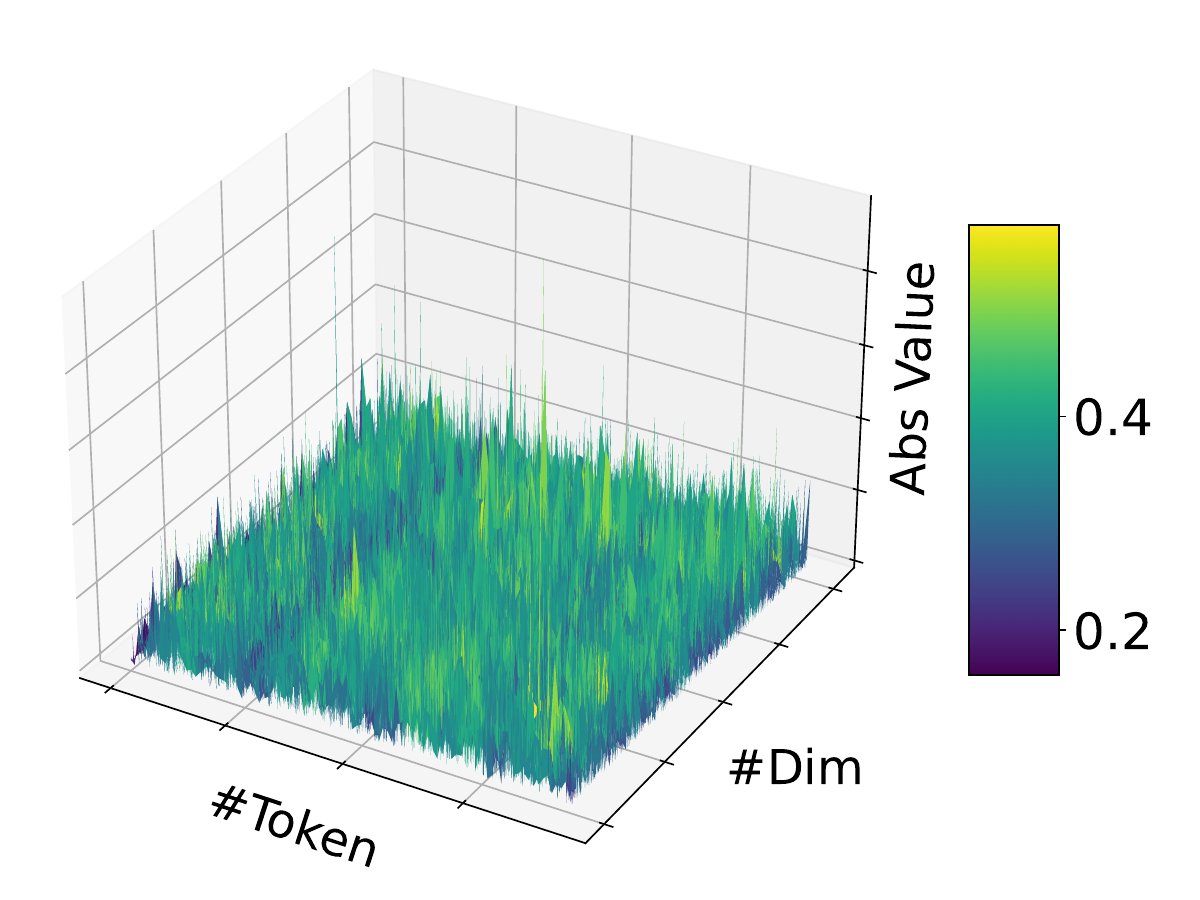}
        \caption{$\mathbf{W}_{\text{qkv}}$ of \our{}}
    \end{subfigure}
    \begin{subfigure}{0.25\textwidth}
        \includegraphics[width=\textwidth]{pics/3b-a8-had.layer.15.o.pdf}
        \caption{$\mathbf{W}_{\text{o}}$ of \our{}}
    \end{subfigure}
    \begin{subfigure}{0.25\textwidth}
        \includegraphics[width=\textwidth]{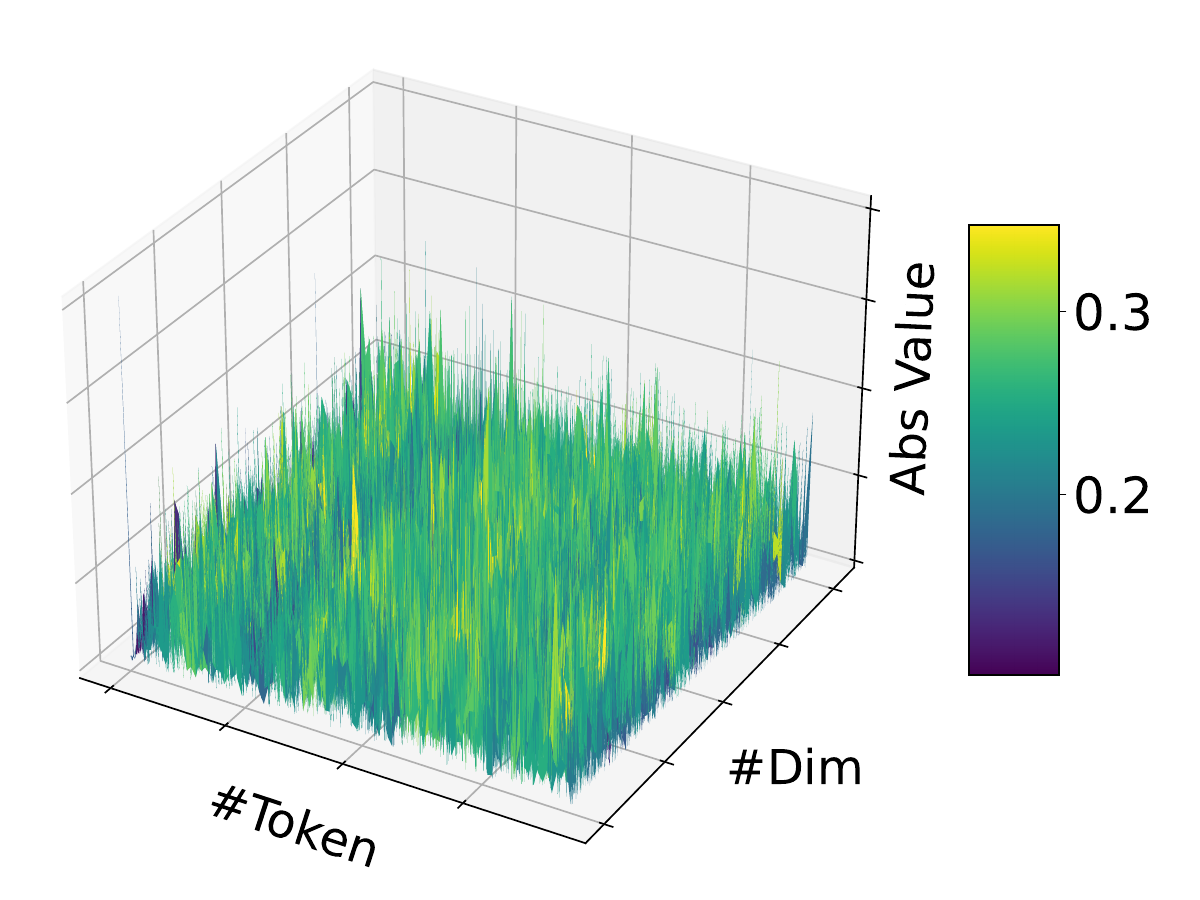}
        \caption{$\mathbf{W}_{\text{up,gate}}$ of \our{}}
    \end{subfigure}
    \begin{subfigure}{0.24\textwidth}
        \includegraphics[width=\textwidth]{pics/3b-a8-had.layer.15.down.pdf}
        \caption{$\mathbf{W}_{\text{down}}$ of \our{}}
    \end{subfigure}
    \caption{The activation distribution of BitNet b1.58 and \our{} with 8-bit activations.}
    \label{fig:distribution_3d}
\end{figure*}

\begin{figure*}[t]
    \centering
    \begin{subfigure}{0.24\textwidth}
        \includegraphics[width=\textwidth]{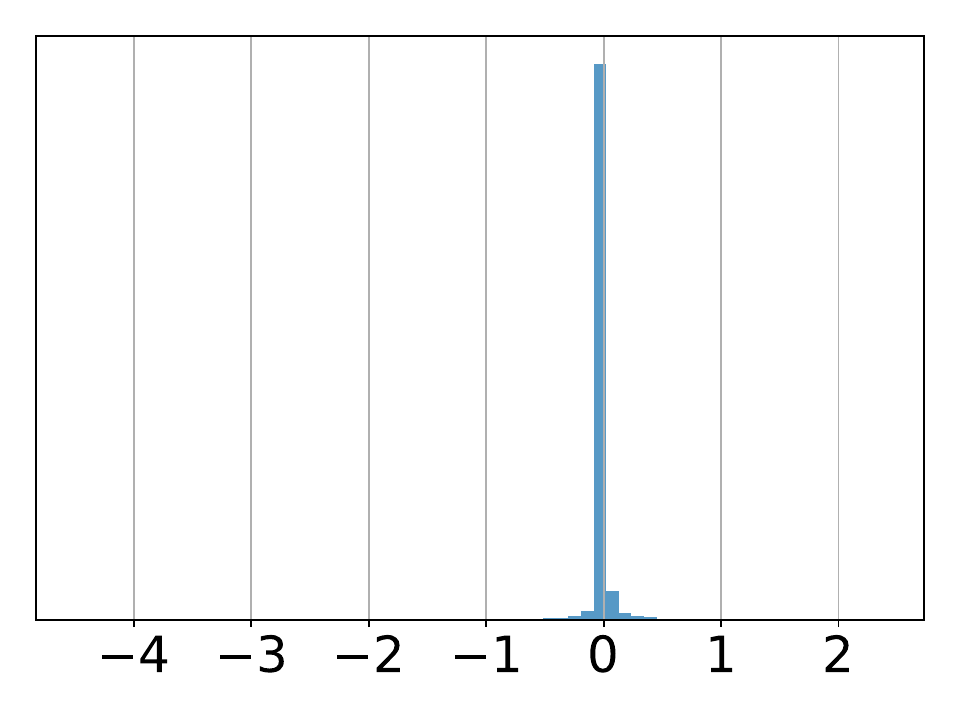}
        \caption{$\mathbf{W}_{\text{down}}$ of BitNet b1.58}
    \end{subfigure}
    \begin{subfigure}{0.24\textwidth}
        \includegraphics[width=\textwidth]{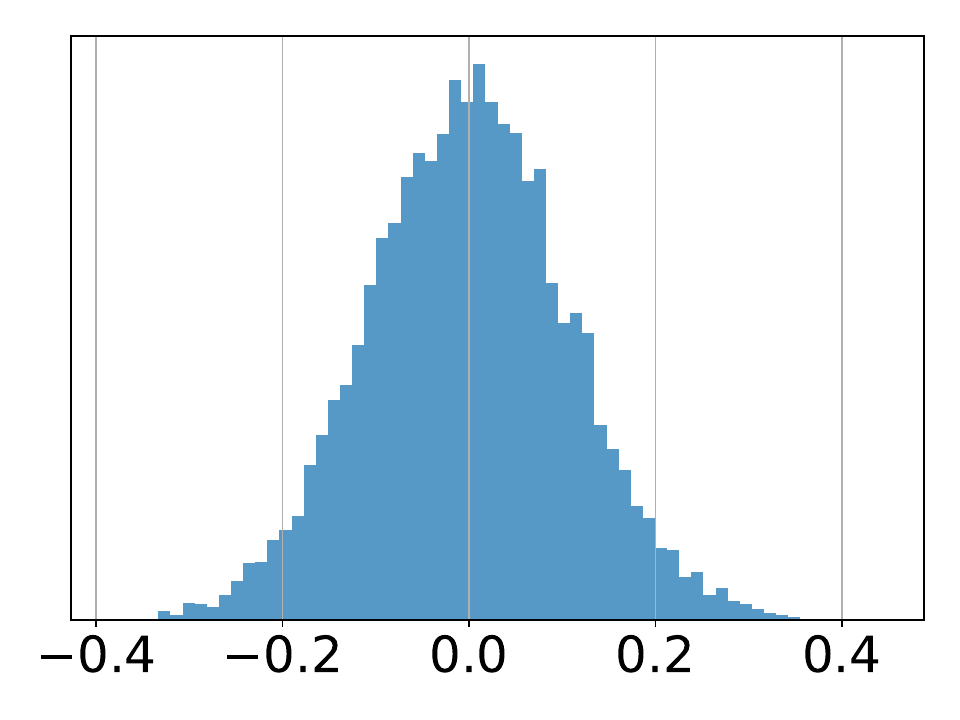}
        \caption{$\mathbf{W}_{\text{down}}$ of \our{}}
    \end{subfigure}
    \begin{subfigure}{0.24\textwidth}
        \includegraphics[width=\textwidth]{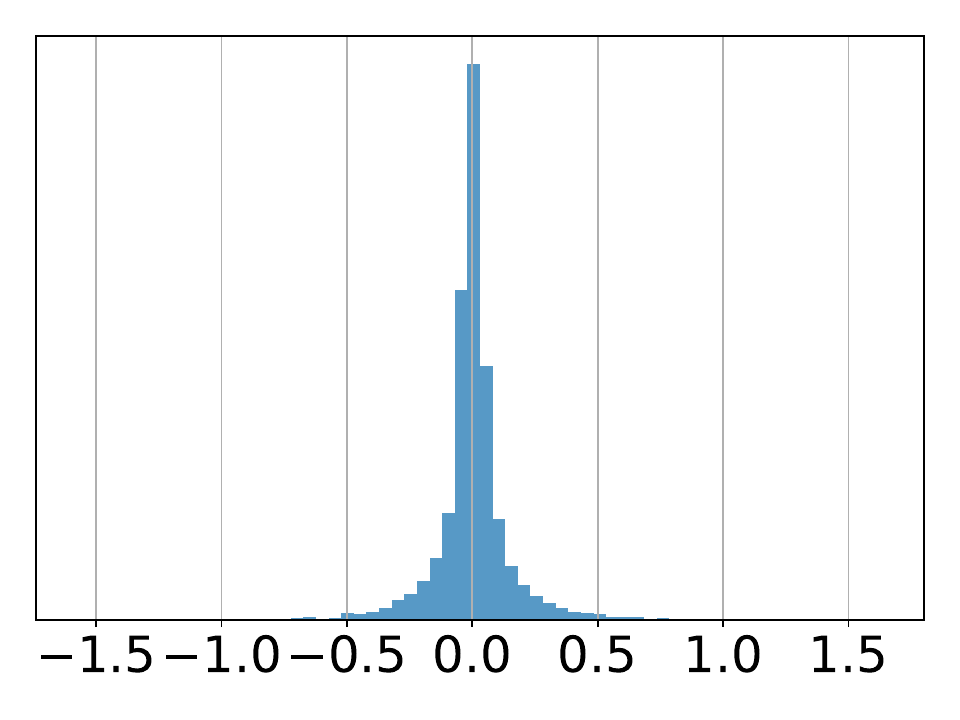}
        \caption{$\mathbf{W}_{\text{o}}$ of BitNet b1.58}
    \end{subfigure}
    \begin{subfigure}{0.24\textwidth}
        \includegraphics[width=\textwidth]{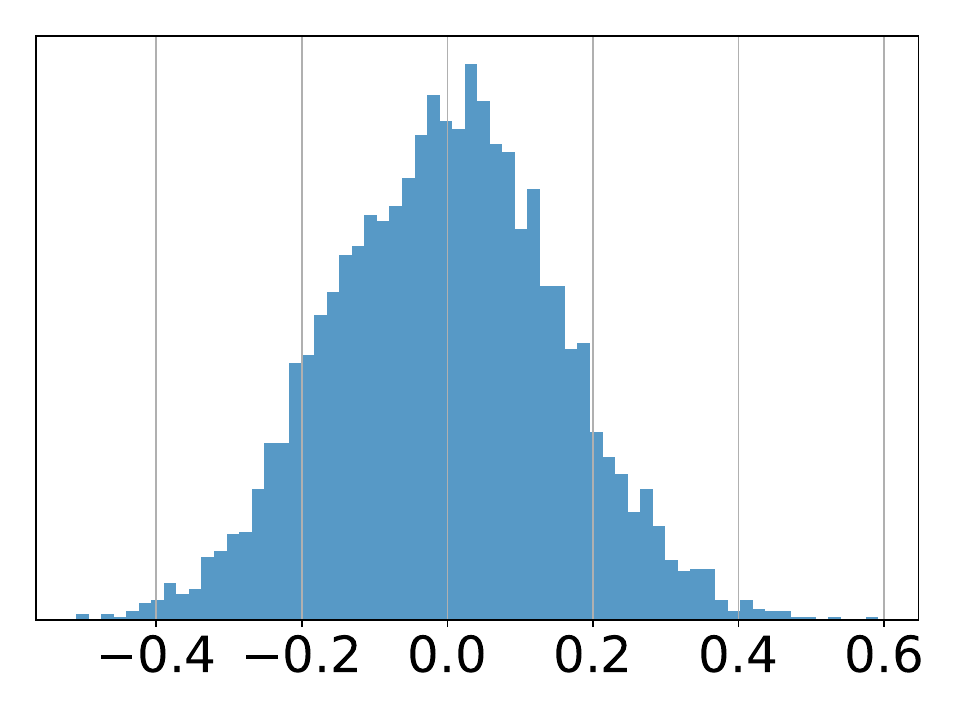}
        \caption{$\mathbf{W}_{\text{o}}$ of \our{}}
    \end{subfigure}
    \caption{The activation distribution of $\mathbf{W}_{\text{down}}$ in FFN and $\mathbf{W}_{\text{o}}$ in attention of BitNet b1.58 and \our{} with 8-bit activations.}
    \label{fig:distribution}
\end{figure*}

As shown in Figure~\ref{fig:distribution_3d} and Figure~\ref{fig:distribution}, with Hadamard transformation, the distribution of the intermediate states becomes closer to a Gaussian-like distribution, which significantly reduces the number of outliers and make it more suitable for INT4 quantization. 

For 8-bit and 4-bit activations, we adopt per-token \textit{absmax} and \textit{absmean} function, respectively. The activation quantization can be formulated as:

\begin{align}
    &\text{Q}_{\text{INT8}}(\mathbf{X}) = \frac{\gamma}{127}\text{RoundClip}(\frac{127}{\gamma+\epsilon}\mathbf{X}, -128, 127),\,\gamma = \max(|\mathbf{X}|) \\
    &\text{Q}_\text{INT4}(\mathbf{X}) = \frac{\beta}{\sqrt{7}}\text{RoundClip}(\frac{\sqrt{7}}{\beta+\epsilon}\mathbf{X}, -8, 7),\,\beta = \text{mean}(|\mathbf{X}|) 
\end{align}

Above all, the matrix multiplication of \rotbitlinear{} can be written as:

\begin{align}
    &\mathbf{Y} = \text{Q}_{w}(\mathbf{W}) \cdot \text{Q}_{\text{INT8/4}}(\mathbf{X_r}),\,\, \mathbf{X_r} = \text{Hadamard}(\text{LN}(\mathbf{X}))
\end{align}

where LN denotes the layer normalization.  

\begin{table*}[t]
    \centering
    \setlength{\tabcolsep}{4.5pt}
    \begin{tabular}{lccccccccc}
    \toprule
    \textbf{Models} & \textbf{Size} & \textbf{PPL$\downarrow$} & \textbf{ARCc$\uparrow$} & \textbf{ARCe$\uparrow$} & \textbf{HS$\uparrow$} & \textbf{PQ$\uparrow$} & \textbf{WGe$\uparrow$} & \textbf{LBA$\uparrow$} & \textbf{Avg$\uparrow$} \\
    \midrule
    \bitnet & \multirow{4}{*}{400M} & 13.37 & 24.32 & 43.01 & 39.51 & 64.91 & 51.93 & 45.51 & 44.87\\ 
    \bitnetx &  & 13.61 & 24.15 & 41.75 & 39.48 & 65.18 & 53.59 & 44.34 & 44.75 \\
    \textbf{\our{} (a8)} & & 13.50 & 23.29 & 43.06 & 39.06 & 64.74 & 50.59 & 45.26 & 44.33\\
    \textbf{\our{} (a4)} & & 13.78 & 23.29 & 41.46 & 38.33 & 65.45 & 50.59 & 44.56 & 43.95 \\
    \midrule
    \bitnet & \multirow{4}{*}{1.3B} & 11.02 & 27.90 & 49.58 & 48.85 & 69.80 & 55.80 & 54.12 & 51.01 \\
    \bitnetx &  & 11.15 & 27.47 & 49.20 & 48.72 & 69.64 & 56.51 & 53.85 & 50.90 \\
    \textbf{\our{} (a8)} & & 11.14 & 27.90 & 49.96 & 48.37 & 69.42 & 57.22 & 54.14 & 51.17 \\
    \textbf{\our{} (a4)} & & 11.33 & 27.56 & 49.58 & 48.00 & 68.23 & 55.49 & 53.58 & 50.41 \\
    \midrule
    \bitnet & \multirow{4}{*}{3B} & 9.71 & 28.84 & 54.80 & 56.39 & 71.44 & 59.35 & 60.47 & 55.22 \\ 
    \bitnetx &  & 9.80 & 29.01 & 55.01 & 55.92 & 71.76 & 59.59 & 59.85 & 55.19 \\
    \textbf{\our{} (a8)} & & 9.72 & 30.55 & 55.56 & 57.19 & 71.33 & 58.72 & 60.90 & 55.71 \\
    \textbf{\our{} (a4)} & & 9.85 & 28.92 & 55.01 & 56.59 & 71.65 & 59.67 & 60.74 & 55.43 \\
    \midrule
    \bitnet & \multirow{4}{*}{7B} & 9.09 & 31.74 & 59.51 & 61.49 & 74.37 & 59.98 & 61.63 & 58.12\\ 
    \bitnetx & & 9.16 & 31.91 & 59.09 & 61.06 & 74.16 & 59.67 & 61.54 & 57.91 \\
    \textbf{\our{} (a8)} & & 9.14 & 32.94 & 58.54 & 61.08 & 74.10 & 61.48 & 64.22 & 58.73 \\
    \textbf{\our{} (a4)} & & 9.24 & 32.42 & 58.00 & 60.71 & 74.27 & 60.85 & 63.52 & 58.30 \\
    \bottomrule
    \end{tabular}
    \caption{Perplexity and results of \our{}, \bitnetx{} and \bitnet{} on the end tasks.}
    \label{tab:zero-shot}
\end{table*}

\subsection{Training}
\label{sec:train}

Following~\cite{bitnet}, we employ the straight-through estimator (STE)~\cite{ste} for gradient approximation and use mixed-precision training to update the parameters. During backward propagation, we bypass the non-differentiable functions in quantization. To support mixed-precision training, we maintain a full-precision latent weight to accumulate parameter updates.

For the backward pass through the Hadamard transformation, we apply the transformation to the gradients as well, leveraging the orthogonality of the transformation matrix $\mathbf{H_m}$. Specifically, the backward propagation is formulated as:

\begin{align}
    \cfrac{\partial \mathcal{L}}{\partial \mathbf{X}} = \text{Hadamard}(\cfrac{\partial \mathcal{L}}{\partial\,\text{Hadamard}(\mathbf{X})})
\end{align}

Similar to BitNet a4.8, \our{} with 4-bit activations can be continue-trained from its 8-bit activation counterpart using a small number of training tokens, while incurring negligible performance loss. The optimizer states are reused for continue-training.

\section{Experiments}
We compared \our{} to BitNet b1.58 and BitNet a4.8 of various model sizes. All models were trained with 1.58-bit weights. BitNet b1.58 has fully INT8 activations for all linear layers. BitNet a4.8 is continue-trained from BitNet b1.58 with a hybrid quantization and sparsification for activations, where the inputs to the sublayers are quantized into 4-bit integers and the intermediate states uses top-$K$ sparsification~\cite{qsparse} and squared ReLU.

We adopted the two-stage weight decay and learning rate scheduling following the training recipe of BitNet b1.58~\cite{bitnet2}. All models were trained with 100B tokens from the RedPajama dataset~\cite{redpajama} to ensure a fair comparison. For \our{} (a4) and BitNet a4.8, we first trained the model with 8-bit activations for 95B tokens. Then we reused the optimizer states and continue-train the model with 4-bit activations for 5B tokens. More details can be found in the Appendix~\ref{ap:hyper}.

We evaluated the zero-shot accuracy for these models on a range of language tasks using the \emph{lm-evaluation-harness} toolkit~\citep{eval-harness}, including ARC-Easy (ARCe)~\cite{arc}, ARC-Challenge (ARCc)~\cite{arc}, Hellaswag (HS)~\cite{hellaswag}, Winogrande (WGe)~\cite{winoGrande} and PIQA (PQ)~\cite{piqa} and LAMBADA (LBA)~\cite{lambada}. We also reported the perplexity on the validation set of C4~\cite{c4} dataset.

\begin{table*}[t]
    \centering
    \begin{tabular}{lcccccccc}
    \toprule
    \textbf{Models} & \textbf{Size} & \textbf{ARCc$\uparrow$} & \textbf{ARCe$\uparrow$} & \textbf{HS$\uparrow$} & \textbf{PQ$\uparrow$} & \textbf{WGe$\uparrow$} & \textbf{LBA$\uparrow$} & \textbf{Avg$\uparrow$} \\
    \midrule
    \our{} (a8) & \multirow{4}{*}{3B} & 30.55 & 55.56 & 57.19 & 71.33 & 58.72 & 60.90 & 55.71 \\
    w/ 4-bit KV & &  29.52 & 55.18 & 57.17 & 70.95 & 58.56 & 60.84 & 55.37\\
    w/ 4-bit QKV & & 30.63 & 55.22 & 57.15 & 71.16 & 58.96 & 60.49 & 55.60\\
    w/ 4-bit Q, 3-bit KV & & 29.69 & 55.22 & 56.22 & 71.49 & 57.62 & 59.01 & 54.88 \\
    \midrule
    \our{} (a8) & \multirow{4}{*}{7B} & 32.94 & 58.54 & 61.08 & 74.10 & 61.48 & 64.22 & 58.73 \\
    w/ 4-bit KV & & 33.02 & 58.67 & 61.04 & 73.61 & 61.88 & 64.06 & 58.71 \\
    w/ 4-bit QKV & & 32.76 & 58.46 & 61.01 & 74.10 & 60.85 & 63.87 & 58.51\\
    w/ 4-bit Q, 3-bit KV & & 32.51 & 58.29 & 60.85 & 73.39 & 60.77 & 62.99 & 58.13\\
    \bottomrule
    \end{tabular}
    \caption{The zero-shot accuracy of \our{} with 8-bit activations and QKV states varying bit-widths on the end tasks.}
    \label{tab:attn_a8}
\end{table*}

\begin{table*}[t]
    \centering
    \begin{tabular}{lcccccccc}
    \toprule
    \textbf{Models} & \textbf{Size} & \textbf{ARCc$\uparrow$} & \textbf{ARCe$\uparrow$} & \textbf{HS$\uparrow$} & \textbf{PQ$\uparrow$} & \textbf{WGe$\uparrow$} & \textbf{LBA$\uparrow$} & \textbf{Avg$\uparrow$} \\
    \midrule
    \our{} (a4) & \multirow{4}{*}{3B} & 28.92 & 55.01 & 56.59 & 71.65 & 59.67 & 60.74 & 55.43 \\
    w/ 4-bit KV & & 29.52 & 54.46 & 56.36 & 71.49 & 58.17 & 60.14 & 55.02 \\
    w/ 4-bit QKV & & 28.58 & 55.43 & 56.32 & 71.16 & 57.93 & 60.70 & 55.02\\
    w/ 4-bit Q, 3-bit KV & & 29.18 & 55.51 & 55.85 & 71.60 & 58.41 & 59.54 & 55.02 \\
    \midrule
    \our{} (a4) & \multirow{4}{*}{7B} & 32.42 & 58.00 & 60.71 & 74.27 & 60.85 & 63.52 & 58.30 \\
    w/ 4-bit KV & & 32.94 & 58.12 & 60.33 & 74.21 & 61.01 & 63.65 & 58.38 \\
    w/ 4-bit QKV & & 33.11 & 57.91 & 60.78 & 74.05 & 61.17 & 62.93 & 58.33\\
    w/ 4-bit Q, 3-bit KV & & 32.08 & 57.95 & 60.29 & 73.23 & 59.59 & 62.97 & 57.69 \\
    \bottomrule
    \end{tabular}
    \caption{The zero-shot accuracy of \our{} with 4-bit activations and QKV states varying bit-widths on the end tasks.}
    \label{tab:attn_a4}
\end{table*}

\subsection{Main Results}
We present the detailed results of \our{} and the baselines in Table~\ref{tab:zero-shot}. Introducing the Hadamard transformation before the quantization in attention and FFN layers results in minimal perplexity degradation. For 8-bit activations, \our{} surpasses BitNet b1.58 with an average accuracy improvement of 0.16\%, 0.49\%, and 0.61\% on end tasks for the 1.3B, 3B, and 7B model sizes, respectively. Additionally, \our{} enables native 4-bit activations across all linear layers, enhancing efficiency for batched inference. With INT4 activations, \our{} achieves perplexity comparable to BitNet a4.8 while demonstrating superior performance on downstream tasks for the 3B and 7B models.

Table~\ref{tab:attn_a8} and Table~\ref{tab:attn_a4} summarize detailed results of \our{} (a8) and \our{} (a4) with low-bit attention, respectively. We adopt post-RoPE quantization for QKV states. The QKV heads were directly quantized to unsigned integers using the absmax function, without the need of any calibration dataset. We retain the KV heads of [BOS] token as 8-bit precision. As shown in Table~\ref{tab:attn_a8} and Table~\ref{tab:attn_a4}, \our{} with 3-bit KV Cache achieves accuracy comparable to its counterpart with full-precision KV cache in 3B and 7B models.

\subsection{Comparison with Post-Training Quantization.} 

We compared \our{} (a4) with post-training quantization baselines, including SpinQuant~\cite{spinquant} and QuaRot~\cite{quarot}, in 1.3B models. QuaRot employs randomized Hadamard transformations to mitigate outlier features, while SpinQuant uses the learnable rotary matrix. Then they adopt GPTQ and \textit{absmax} function to quantize the weight and activations into 4-bit, respectively. Since the weights of BitNet b1.58 were already trained to be ternary values from scratch, we adopted \textit{absmean} function used in the training of BitNet rather than GPTQ for weight quantization. For activation quantization of the baselines, we retained rotary transformations for activations as an online operation; Furthermore, we removed the fusion of RMS normalization scales into projections. Following~\cite{spinquant}, we tuned the rotation of SpinQuant with 800 samples from WikiText2 dataset.

We report the results of \our{} (a4) and BitNet b1.58 with SpinQuant and QuaRot in Table~\ref{tab:ptq}. All models were quantized to 1.58-bit weight and 4-bit activations. \our{} (a4) significantly surpasses these baselines in terms of perplexity on valid set of C4 and accuracy on downstream tasks. Additionally, we observed that ternary models are more sensitive to the fusion of rotary matrix and latent weights. Specifically, we removed rotary matrix fusion for $\text{W}_{\text{qkv}}$ in attention modules and $\text{W}_{\text{up, gate}}$ in FFNs, i.e., using $\text{Q}_\text{w}(W)R^T$ rather than $\text{Q}_\text{w}(WR^T)$. Removing this fusion notably enhances baseline performance: the perplexity decreases from 19.80 to 13.52 for SpinQuant. However, they still trails substantially behind \our{} (a4). Moreover, this adjustment forces these projections to revert to full precision (W16A4), thus sacrificing inference efficiency.

\begin{table*}[t]
    \setlength{\tabcolsep}{7pt}
    \centering
    \begin{tabular}{lcccccccc}
    \toprule
    \textbf{Models} & \textbf{PPL$\downarrow$} & \textbf{ARCc$\uparrow$} & \textbf{ARCe$\uparrow$} & \textbf{HS$\uparrow$} & \textbf{PQ$\uparrow$} & \textbf{WGe$\uparrow$} & \textbf{LBA$\uparrow$} & \textbf{Avg$\uparrow$} \\
    \midrule
    \multicolumn{6}{l}{\ \emph{w/o fusing rotary matrix to $\text{W}_\text{qkv,up,gate}$}} \\
    \color{gray!70} QuaRot & \color{gray!70} 13.52 & \color{gray!70} 26.28 & \color{gray!70} 47.43 & \color{gray!70} 45.92 & \color{gray!70} 65.89 & \color{gray!70} 51.46 & \color{gray!70} 42.34 & \color{gray!70} 46.55\\
    \color{gray!70} SpinQuant & \color{gray!70} 13.52 & \color{gray!70} 25.60 & \color{gray!70} 47.35 & \color{gray!70} 45.52 & \color{gray!70} 67.25 & \color{gray!70} 52.49 & \color{gray!70} 42.52 & \color{gray!70} 46.79 \\
    \midrule
    QuaRot & 20.83 & 24.74 & 40.78 & 40.54 & 62.89 & 49.33 & 36.89 & 42.53 \\
    SpinQuant & 19.80 & 24.74 & 40.19 & 40.77 & 62.73 & 52.09  & 39.24 & 43.29 \\
    \bf \our{} (a4) & \bf 11.33 & \bf 27.56 & \bf 49.58 & \bf 48.00 & \bf 68.23 & \bf 55.49 & \bf 53.58 & \bf 50.41 \\
    \bottomrule
    \end{tabular}
    \caption{Perplexity and zero-shot accuracy of \our{}, QuaRot and SpinQuant on the end tasks.}
    \label{tab:ptq}
\end{table*}

\subsection{Ablation Study}

We conduct ablation studies on the Hadamard transformation in \rotbitlinear{} at the 1.3B and 3B model scales. All models are trained on the same dataset to ensure a fair comparison. We report perplexity on the C4 validation set and zero-shot accuracy across a range of language tasks, including ARC-Easy~\cite{arc}, ARC-Challenge~\cite{arc}, Hellaswag~\cite{hellaswag}, Winogrande~\cite{winoGrande}, PIQA~\cite{piqa}, and LAMBADA~\cite{lambada}. As shown in Table~\ref{tab:ablation}, removing the rotary transformation leads to model divergence. Moreover, while applying the Hadamard transformation to both the weights and activations results in faster convergence, it achieves similar performance to applying it only to the activations as training progresses. Therefore, for simplicity, we apply the Hadamard transformation only to the activations in \rotbitlinear{}. Detailed results can be found in Appendix~\ref{ap:res}.

\begin{table*}[t]
    \centering
    \begin{tabular}{l|c|cc|cc}
    \toprule
    \multirow{2}{*}{\textbf{Methods}} & \multirow{2}{*}{\textbf{\#Bits}} & \multicolumn{2}{c}{\textbf{1.3B}} & \multicolumn{2}{c}{\textbf{3B}} \\
    & & Acc.$\uparrow$ & PPL$\downarrow$ & Acc.$\uparrow$ & PPL$\downarrow$\\
    \midrule
    No rotation & \multirow{3}{*}{W1.58A8} & 51.01 & 11.02 & 55.22 & 9.71 \\ 
    Weight \& activation rotation &  & 50.47 & 11.14 & 55.55 & 9.69 \\
    Activation rotation & & 51.16 & 11.14 & 55.71 & 9.72 \\
    \midrule
     No rotation & \multirow{3}{*}{W1.58A4} & \multicolumn{2}{c|}{diverged} & \multicolumn{2}{c}{diverged} \\
    Weight \& activation rotation &  & 50.09 & 11.33 & 54.98 & 9.81 \\
    Activation rotation & & 50.41 & 11.33 & 55.43 & 9.85\\
    \bottomrule
    \end{tabular}
    \caption{Ablations on the Hadamard transformation of \rotbitlinear{} across various sizes.}
    \label{tab:ablation}
\end{table*}

\section{Conclusion}

We introduce \our{}, enabling native 4-bit activations within 1-bit LLMs. This is achieved using our proposed \rotbitlinear{} layer in place of standard attention output and FFN down projections. \rotbitlinear{} employs an online Hadamard transformation before activation quantization to effectively suppress outlier channels by reshaping the activation distribution. Our experiments show \our{} with 8-bit activations matches BitNet b1.58 performance. Subsequently, \our{} can be trained for native 4-bit activation use. This 4-bit variant, \our{}(a4), maintains comparable performance to the 8-bit version while significantly boosting efficiency in batched inference scenarios.

\bibliography{bitnet}
\bibliographystyle{alpha}

\appendix

\section{More results}
\label{ap:res}

\begin{table*}[h]
    \centering
    \setlength{\tabcolsep}{4.5pt}
    \begin{tabular}{lcccccccc}
    \toprule
    \textbf{Methods} & \textbf{Size} & \textbf{ARCc$\uparrow$} & \textbf{ARCe$\uparrow$} & \textbf{HS$\uparrow$} & \textbf{PQ$\uparrow$} & \textbf{WGe$\uparrow$} & \textbf{LBA$\uparrow$} & \textbf{Avg$\uparrow$} \\
    \midrule
    No rotation & \multirow{3}{*}{1.3B} & \multicolumn{7}{c}{diverged} \\
    Weight \& activation rotation & & 27.13 & 50.29 & 48.32 & 69.04 & 54.30 & 53.76 & 50.47\\
    Activation rotation & & 27.90 & 49.96 & 48.37 & 69.42 & 57.22 & 54.14 & 51.17\\
    \midrule
    No rotation & \multirow{3}{*}{3B} & \multicolumn{7}{c}{diverged} \\
    Weight \& activation rotation & & 30.03 & 55.72 & 56.81 & 71.65 & 59.43 & 59.65 & 55.54 \\
    Activation rotation & & 30.55 & 55.56 & 57.19 & 71.33 & 58.72 & 60.90 & 55.71\\
    \bottomrule
    \end{tabular}
    \caption{Ablations on the Hadamard transformation of \rotbitlinear{} across various sizes. All models have 1.58-bit weights and 8-bit activations.}
\end{table*}

\begin{table*}[h]
    \centering
    \setlength{\tabcolsep}{4.5pt}
    \begin{tabular}{lcccccccc}
    \toprule
    \textbf{Methods} & \textbf{Size} & \textbf{ARCc$\uparrow$} & \textbf{ARCe$\uparrow$} & \textbf{HS$\uparrow$} & \textbf{PQ$\uparrow$} & \textbf{WGe$\uparrow$} & \textbf{LBA$\uparrow$} & \textbf{Avg$\uparrow$} \\
    \midrule
    No rotation & \multirow{3}{*}{1.3B} & \multicolumn{7}{c}{diverged} \\
    Weight \& activation rotation & & 27.22 & 49.12 & 47.77 & 69.37 & 54.54 & 52.49 & 50.09\\
    Activation rotation & & 27.56 & 49.58 & 48.00 & 68.23 & 55.49 & 53.58 & 50.41\\
    \midrule
    No rotation & \multirow{3}{*}{3B} & \multicolumn{7}{c}{diverged} \\
    Weight \& activation rotation & & 29.44 & 54.46 & 56.57 & 71.93 & 57.85 & 59.64 & 54.98 \\
    Activation rotation & & 28.92 & 55.01 & 56.59 & 71.65 & 59.67 & 60.74 & 55.43\\
    \bottomrule
    \end{tabular}
    \caption{Ablations on the Hadamard transformation of \rotbitlinear{} across various sizes. All models have 1.58-bit weights and 4-bit activations.}
\end{table*}

\section{Hyper-parameters}
\label{ap:hyper}

\begin{table*}[h]
\setlength{\tabcolsep}{5pt}
\centering
\begin{tabular}{cccccccc}
\toprule
\bf Size & \bf Hidden Size & \bf GLU Size & \bf \#Heads & \bf \#Layers & \bf Batch Size & \bf \# Tokens & \bf Seq Length \\
\midrule
400M & 1024 & 4096 & 16 & 24 & 1M & 100B & 2048 \\
1.3B & 2048 & 8192 & 32 & 18 & 1M & 100B  & 2048  \\
3B & 4096 & 8192 & 32 & 20 & 1M & 100B  & 2048  \\
7B & 4096 & 16384 & 32 & 24 & 1M & 100B  & 2048  \\
\bottomrule
\end{tabular}
\caption{Model configurations for the BitNet models.}
\end{table*}

\begin{table*}[h]
\centering
\begin{tabular}{lcccccc}
\toprule
\bf Model & \bf Size & \bf Learning Rate & \bf Weight Decay & \bf Warm-up & \bf Adam $\beta$ \\
\midrule
\multirow{4}{*}{BitNet} & 400M & $1.8\times10^{-3} \rightarrow 1.2\times10^{-3}$ & $0.1 \rightarrow 0$ & 375 & (0.9, 0.95) \\
& 1.3B & $1.2\times10^{-3} \rightarrow 8\times10^{-4}$ & $0.1 \rightarrow 0$ & 375 & (0.9, 0.95)  \\
& 3B &  $1.2\times10^{-3} \rightarrow 6.4\times10^{-4}$ & $0.1 \rightarrow 0$ & 375 & (0.9, 0.95) \\
& 7B &  $1\times10^{-3} \rightarrow 6\times10^{-4}$ & $0.1 \rightarrow 0$ & 375 & (0.9, 0.95) \\
\bottomrule
\end{tabular}
\caption{Hyper-parameters for both \our{} training.}
\end{table*}

\end{document}